\documentclass[runningheads]{llncs}
\usepackage[T1]{fontenc}
\usepackage{graphicx}
\PassOptionsToPackage{table, dvipsnames, svgnames, x11names}{xcolor}

\usepackage{colortbl}       
\usepackage{booktabs}
\usepackage{multirow}
\usepackage{tikz}
\usetikzlibrary{shadows}
\usetikzlibrary{shadows.blur, backgrounds}
\usepackage{booktabs}
\usepackage{multirow}
\usepackage{tcolorbox}
    \tcbuselibrary{skins}
\newtcolorbox{mybox}[1]{
    colback=brown!5!white,
    colframe=brown!5!black,
    borderline={2pt}{0mm}{black},
    borderline={.7pt}{1mm}{black},
    fonttitle=\bfseries,
    title=#1,
    arc=3mm,
    segmentation hidden,
    top=8pt,
    bottom=8pt
}

\usepackage{latexsym}
\usepackage{enumitem}
\usepackage{graphicx}

\usepackage[
  colorlinks=true,
  linkcolor=blue,
  citecolor=blue,
  urlcolor=blue
]{hyperref}

\usepackage{orcidlink}

\begin{document}
%
\title{Evaluating Vision-Language and Large Language Models for Automated Student Assessment in Indonesian Classrooms}
\titlerunning{Evaluating VL and LLMs for Automated Student Assessment}
%
\author{Nurul Aisyah\inst{1}\orcidlink{0009-0003-3500-831X} \and
Muhammad Dehan Al Kautsar\inst{2}\orcidlink{0009-0000-5416-3524} \and
Arif Hidayat\inst{3}\orcidlink{0000-0003-0734-0756} \and \\
Raqib Chowdhury\inst{4}\orcidlink{0000-0002-3136-6545}
\and
Fajri Koto\inst{2}\orcidlink{0000-0002-3659-6761}
}

\authorrunning{N. Aisyah et al.}
%
\institute{Quantic School of Business and Technology, Washington, DC, USA 
 \and
Mohamed bin Zayed University of Artificial Intelligence, Abu Dhabi, UAE \and
Indonesia University of Education, Bandung, Indonesia \and
Monash University, Melbourne, Australia
\\
\email{na57@students.quantic.edu, fajri.koto@mbzuai.ac.ae}
}
\maketitle              
\begin{abstract}
Despite rapid progress in vision–language and large language models (VLMs and LLMs), their effectiveness for AI-driven educational assessment in real-world, underrepresented classrooms remains largely unexplored.
We evaluate state-of-the-art VLMs and LLMs on over 14K handwritten answers from grade-4 classrooms in Indonesia, covering Mathematics and English aligned with the local national curriculum. Unlike prior work on clean digital text, our dataset features naturally curly, diverse handwriting from real classrooms, posing realistic visual and linguistic challenges. Assessment tasks include grading and generating personalized Indonesian feedback guided by rubric-based evaluation. Results show that the VLM struggles with handwriting recognition, causing error propagation in LLM grading, yet LLM feedback remains pedagogically useful despite imperfect visual inputs, revealing limits in personalization and contextual relevance.\footnote{Data access is granted after completing the form: \url{https://tinyurl.com/4975895v}}

\keywords{AI-driven educational assessment \and Indonesia \and LLMs \and VLMs}
\end{abstract}
\section{Introduction}

Vision–language models (VLMs)~\cite{liu2023visual,liu2024vmamba,steiner2024paligemma} and large language models (LLMs)~\cite{touvron2023llamaopenefficientfoundation} have demonstrated impressive reasoning capabilities \cite{wang2023can,wei2022chain}, including solving complex academic tasks such as university-level physics \cite{yeadon2024impact} and competition-grade mathematics problems \cite{zhang2024accessing}. These advancements have driven growing interest in applying such models to education. Common areas of application include automated grading \cite{chiang2024large}, teaching support \cite{hu2025exploring}, feedback generation \cite{morris2023using}, and content creation \cite{westerlund2024llm}.

\begin{figure}[t]
    \centering
    \includegraphics[width=0.7\linewidth]{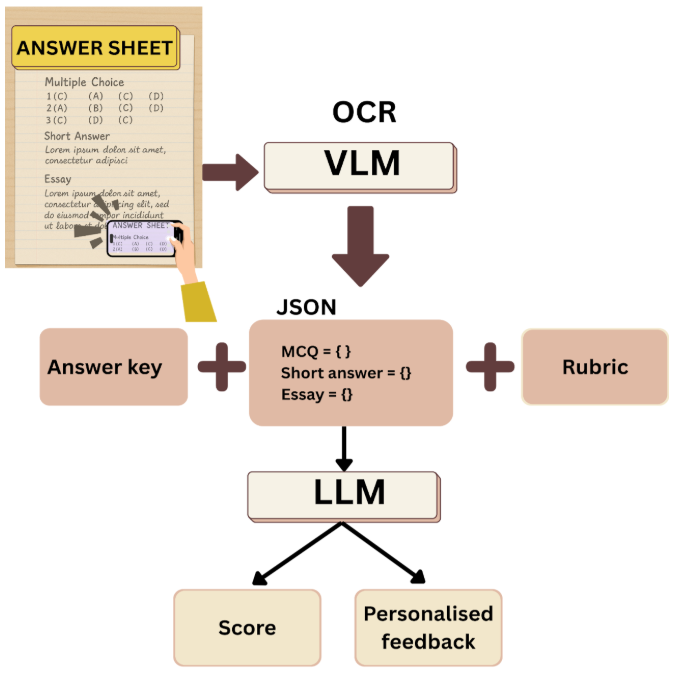}
    \caption{AI-powered assessment using VLM and LLM.}
    \label{fig:question_example}
\end{figure}

However, most VLM- and LLM-based educational tools have been developed in Global North contexts \cite{lee2025realizing,yancey2023rating}, limiting their applicability in classrooms where curricula, languages, and instructional practices differ substantially. Even within these well-resourced settings, only a few studies (e.g., \cite{daniela2025evaluating}) have explored end-to-end AI assessment on handwritten student work—and typically on very small samples, such as 24 responses—while the majority of existing benchmarks continue to rely on clean, digital text \cite{chiang2024large}.

In Indonesia, paper-based assignments remain the preferred practice in most schools. Many schools, especially in rural areas, lack consistent access to digital devices, highlighting the need for AI systems that function effectively in low-tech environments. Second, using handwritten responses helps reduce the risk of academic dishonesty, such as students who rely on AI tools to generate answers. 


Our contributions are as follows:
\begin{enumerate}[topsep=0px,partopsep=0px]
\item We release the first dataset of handwritten student answer sheets for AI-based educational assessment, comprising over 14K responses collected from six primary schools in Indonesia—three rural and three urban. The dataset covers Grade 4 Mathematics and English, with questions and scoring rubrics designed by experienced teachers. All answers were manually transcribed and graded by professional educators.\footnote{To ensure ethical use and protect student privacy, all personally identifiable information (e.g., student names, grade levels, and school names) has been removed.} In Indonesia, English is taught as a foreign language, while Mathematics often involves handwritten explanations in Indonesian 
\item We introduce a multimodal pipeline that integrates vision–language models (VLMs) and large language models (LLMs), as illustrated in Figure~\ref{fig:question_example}. We compare several state-of-the-art models for grading student answers and find that GPT-4o with vision input achieves the highest accuracy and feedback quality.
\item We conduct a manual evaluation of LLM-generated feedback in Indonesian and find that, even when based on imperfect input (e.g., OCR errors), the feedback tends to be clear and factually correct. However, personalization and helpfulness remain notable areas of concern.
\end{enumerate}


\section{Related Work}


Previous studies have investigated the use of LLMs as graders for student assignments and exams. For example, Chiang et al.~\cite{chiang2024large} used GPT-4 to automatically grade 1,028 student essays in a university-level course titled \textit{Introduction to Generative AI}. 
Their findings suggest that LLM-based graders were generally well accepted by students; however, the models occasionally did not follow the grading rubric.
In a related study, Yancey et al. \cite{yancey2023rating} used GPT-3.5 and GPT-4 to score essays in a high-stakes English proficiency test, demonstrating that LLM-generated scores can achieve high agreement with human raters.

Stahl et al. \cite{stahl2024exploring} used Mistral \cite{jiang2023mistral7b} and LLaMA-2 \cite{touvron2023llama} to assess English student essays and generate feedback, finding that scoring accuracy had a limited influence on students’ perceived usefulness of the feedback. Similarly, Morris et al.~\cite{morris2023using} applied a Longformer-based language model~\cite{botarleanu2022multitask}, trained on data primarily collected in the Global North (with 52\% of the data from White individuals), to generate formative feedback on student-written summaries of English textbooks.

Unlike these prior studies, our work focuses on handwritten responses from grade 4 primary school students in Indonesia, covering both English and mathematics. We also evaluate a complete multimodal pipeline that integrates a VLM for handwriting recognition and LLMs for grading and feedback generation—introducing new challenges related to noisy input, multilingual content, and real-world constraints in low-tech, underrepresented classroom settings.

\section{Dataset Construction}
\label{sec:dataset}

\paragraph{Assessment Design}
 We developed assessment instruments for grade 4 primary school students in two subjects: Mathematics and English. The items were designed from scratch based on a thorough analysis of the national curriculum and corresponding learning objectives. Each subject assessment consisted of 10 multiple-choice questions (MCQs), 10 short-answer questions, and 2 essay questions. All items were created by experienced senior subject teachers—an English teacher and a Math teacher—each with over 10 years of classroom experience and a Master’s degree in Education. In addition to writing the assessment items, these teachers developed detailed scoring rubrics for the short-answer and essay questions, as well as answer keys for the MCQs. Standardized answer sheets were also prepared to collect student responses. 

\paragraph{Data Collection} 

Data collection was carried out in six primary schools, evenly divided between rural (Sumatra and Nusa Tenggara Islands) and urban (Java Island) settings. Each classroom included approximately 20 to 30 students. 
Students had up to 30 minutes to complete their answers on a standardized answer sheet.

In total, we collected 646 handwritten answer sheets from these assessments, with a total of more than 14K student answers.
Of these, 414 were collected from urban schools and 232 from rural schools 
The disparity in sample size between urban and rural areas is primarily due to larger class sizes typically found in urban schools compared to their rural counterparts. 

\begin{table}[t]

\centering
\caption{Mean absolute error (MAE) for scoring English and Math, calculated separately for multiple-choice (M), short-answer (S), essay (E), and the total score. Lower values indicate better performance; bolded numbers represent the best results. Scores range from 0 to 100.}
\resizebox{0.95\linewidth}{!}{%
\begin{tabular}{@{}lcccclcccc@{}}
\toprule
    \multicolumn{1}{l}{}     & \multicolumn{4}{c}{\textbf{English}}          & \textbf{} & \multicolumn{4}{c}{\textbf{Math}} \\
    \cmidrule{2-5}\cmidrule{7-10}
    \multicolumn{1}{c}{\multirow{-2}{*}{\textbf{Model}}} & \multicolumn{1}{c}{\textbf{MCQ}}   & \multicolumn{1}{c}{\textbf{Short-Essay}}  & \multicolumn{1}{c}{\textbf{Essay}}   & \multicolumn{1}{c}{\textbf{Total}} & \textbf{} & \multicolumn{1}{l}{\textbf{MCQ}}   & \multicolumn{1}{c}{\textbf{Short-Essay}}  & \multicolumn{1}{c}{\textbf{Essay}}   & \multicolumn{1}{c}{\textbf{Total}} \\
    \midrule
    \rowcolor[HTML]{D9EAD3} 
    
    OCR by GPT4o & \multicolumn{1}{l}{\cellcolor[HTML]{D9EAD3}} & \multicolumn{1}{l}{\cellcolor[HTML]{D9EAD3}} & \multicolumn{1}{l}{\cellcolor[HTML]{D9EAD3}} & \multicolumn{1}{l}{\cellcolor[HTML]{D9EAD3}} & & \multicolumn{1}{l}{\cellcolor[HTML]{D9EAD3}} & \multicolumn{1}{l}{\cellcolor[HTML]{D9EAD3}} & \multicolumn{1}{l}{\cellcolor[HTML]{D9EAD3}} & \multicolumn{1}{l}{\cellcolor[HTML]{D9EAD3}} \\
    {GPT4o} & \textbf{2.8}  & 14.6 & \textbf{5.6}  & \textbf{11.7} & & \textbf{2.3}  & 16.3 & \textbf{1.5}  & 8.2  \\
    {Llama 3.1 (70B)} & \textbf{2.8}  & 18.7 & 9.3  & 14.5 & & \textbf{2.3}  & \textbf{10.6} & 27.5 & \textbf{2.2}  \\
    {Qwen2.5 (72B)}   & \textbf{2.8}  & 14.9 & 16.6 & 14.7 & & \textbf{2.3}  & 19.1 & 5.8  & 7.1  \\
    {Deepseek (671B)} & \textbf{2.8}  & \textbf{12.6} & 9.8  & 11.9 & & \textbf{2.3}  & 22.8 & 6.7  & 8.1  \\
    \rowcolor[HTML]{D9EAD3} 
    OCR by Human & \multicolumn{1}{l}{\cellcolor[HTML]{D9EAD3}} & \multicolumn{1}{l}{\cellcolor[HTML]{D9EAD3}} & \multicolumn{1}{l}{\cellcolor[HTML]{D9EAD3}} & \multicolumn{1}{l}{\cellcolor[HTML]{D9EAD3}} & & \multicolumn{1}{l}{\cellcolor[HTML]{D9EAD3}} & \multicolumn{1}{l}{\cellcolor[HTML]{D9EAD3}} & \multicolumn{1}{l}{\cellcolor[HTML]{D9EAD3}} & \multicolumn{1}{l}{\cellcolor[HTML]{D9EAD3}} \\
    {GPT4o} & \textbf{0.0}    & 9.2  & 2.7  & 7.9  & & \textbf{0.0}  & \textbf{2.9}  & \textbf{5.7}  & 1.5  \\
    {Llama 3.1 (70B)} & \textbf{0.0}   & 14.4 & 2.3  & 11.6 & & \textbf{0.0}  & 9.8  & 19.1 & 10.3 \\
    {Qwen2.5 (72B)}   & \textbf{0.0}    & 8.4  & 3.8  & 9.2  & & \textbf{0.0}  & 5.5  & 8.7  & 3.3  \\
    {Deepseek (671B)} & \textbf{0.0}    & \textbf{4.4}  & \textbf{1.5}  & \textbf{6.8}  & & \textbf{0.0}  & 5.9  & 8.5  & \textbf{0.8} \\
    \bottomrule
\end{tabular}
}

\label{tab:result1}
\end{table}

\section{Experiment}

\paragraph{Overall Pipeline} Figure~\ref{fig:question_example} illustrates our pipeline, which begins with a vision language model (VLM) that performs optical character recognition (OCR) to extract handwritten student responses from scanned answer sheets. The extracted text is then structured into a JSON format and passed to a large language model (LLM), along with the answer key and a teacher-defined rubric. For multiple-choice questions, we apply string matching. For short-answer and essay questions, we run the LLM separately for each question, providing the student’s response, the corresponding answer key, and the assessment rubric. To generate personalized feedback, we provide the LLM with all of the student’s responses, the answer key, the assigned weights, and the rubric.

\paragraph{Model} For OCR, we employ GPT-4o\footnote{\url{https://openai.com/}}, alongside a gold-standard transcription manually prepared by teachers for comaprison. We also evaluated Qwen-VL-72B 
but found its performance suboptimal; hence, GPT-4o was used for all subsequent experiments.
For automatic scoring, we compare the performance of GPT-4o, Llama-3.1-Instruct (70B) \cite{touvron2023llama}, Qwen2.5-Instruct (72B) \cite{qwen2.5}, and DeepSeek-Chat (671B) \cite{liu2024deepseek}.
For personalized feedback generation, we rely on the grading results from GPT-4o, producing two feedback variants using GPT-4o and DeepSeek-Chat, respectively.

\paragraph{Evaluation} Each answer sheet image was manually transcribed and scored by professional teachers. We compared the LLM-generated scores against these gold-standard scores across three question types: multiple-choice, short-answer, and essay, using mean absolute error (MAE) as the evaluation metric. For personalized feedback, we conducted a manual evaluation covering four aspects: Correctness, Personalization, Clarity, and Educational Value/Helpfulness, rated on a 1–5 scale, where 1 indicates the lowest quality.\footnote{This evaluation was carried out by experienced educators with a Master’s degree in teaching.}

\begin{table}[t]
\caption{Human evaluation by expert teachers on personalized feedback, using a rating scale from 1 to 5, where 1 indicates the lowest score.}
\centering
\resizebox{0.8\linewidth}{!}{%
    \begin{tabular}{lcccc}
    \toprule
    \textbf{Model} & \multicolumn{1}{c}{\textbf{Correctness}} & \multicolumn{1}{c}{\textbf{Personalization}} & \multicolumn{1}{c}{\textbf{Clarity}}     & \multicolumn{1}{c}{\textbf{Helpfulness}} \\
    \midrule
    \rowcolor[HTML]{D9EAD3} 
    English    & \multicolumn{1}{c}{\cellcolor[HTML]{D9EAD3}} & \multicolumn{1}{c}{\cellcolor[HTML]{D9EAD3}} & \multicolumn{1}{c}{\cellcolor[HTML]{D9EAD3}} & \multicolumn{1}{c}{\cellcolor[HTML]{D9EAD3}} \\
    GPT-4o     & \textbf{4.00} \textsubscript{($\pm$1.0)}       & \textbf{3.96} \textsubscript{($\pm$0.7)}        & 3.64 \textsubscript{($\pm$0.8)} & 3.60 \textsubscript{($\pm$0.8)}  \\
    Deepseek   & 3.96 \textsubscript{($\pm$0.9)} & 3.88 \textsubscript{($\pm$0.8)} & \textbf{4.04} \textsubscript{($\pm$0.9)}        & \textbf{3.96} \textsubscript{($\pm$1.1)}        \\
    \rowcolor[HTML]{D9EAD3} 
    Math   & \multicolumn{1}{l}{\cellcolor[HTML]{D9EAD3}} & \multicolumn{1}{l}{\cellcolor[HTML]{D9EAD3}} & \multicolumn{1}{l}{\cellcolor[HTML]{D9EAD3}} & \multicolumn{1}{l}{\cellcolor[HTML]{D9EAD3}} \\
    GPT-4o     & 3.84 \textsubscript{($\pm$0.9)} & \textbf{3.72} \textsubscript{($\pm$0.9)}        & 3.92 \textsubscript{($\pm$1.2)} & \textbf{3.68} \textsubscript{($\pm$0.8)}        \\
    Deepseek   & \textbf{3.88} \textsubscript{($\pm$0.3)}        & 2.96 \textsubscript{($\pm$0.5)} & \textbf{4.00} \textsubscript{($\pm$0.0)}       & 2.92 \textsubscript{($\pm$0.3)}    \\
    \bottomrule
    \end{tabular}
}

\label{tab:analysis}
\end{table}

\section{Result and Analysis}

\paragraph{Main Result}
Table~\ref{tab:result1} presents the performance of the LLMs selected in three types of questions: multiple choice, short answer, and essay. When using GPT-4o to extract student responses via OCR, we observe that most model-generated scores are generally competitive. Among them, GPT-4o produces scores that align most closely with human grading for essay questions, achieving the lowest MAE in both English (5.6) and Math (1.5). In contrast, LLaMA-3.1–70B and Qwen-2.5–72B are less reliable, with scores deviating more significantly from human judgments. Short-answer questions remain the most challenging to evaluate: even the best-performing model in this category, LLaMA-3.1-7B for Math, still shows a relatively high MAE of 10.6, indicating a notable gap from human-level accuracy.

However, the results differ when human effort is involved in the OCR task. Most scores become better overall, with Deepseek-chat and GPT-4o emerging as the top-performing models. Deepseek-chat shows strong performance in English (MAE of 4.4 for short answers and 1.5 for essays), while GPT-4o performs best in Math, with only a 2.9 difference in short answers and 5.7 in essays. It is worth noting that MCQ scores remain at 0, as basic string matching is sufficient due to the exact nature of the answers. 

\paragraph{Human Evaluation on Personalised Feedback}
Table~\ref{tab:analysis} presents the results of a human evaluation on personalized feedback quality, rated by two expert teachers across four dimensions: Correctness, Personalization, Clarity, and Helpfulness (scale 1–5, with scores below 3 considered poor). Inter-annotator agreement was measured using Spearman correlation, yielding values between 0.4 and 0.7 across all dimensions, which indicates moderate to strong agreement between the two evaluators. For English, GPT-4o slightly outperforms Deepseek in correctness and personalization, while Deepseek leads in clarity and helpfulness. In Math, Deepseek shows strong clarity and correctness but performs poorly in personalization and helpfulness, with both scores falling below 3. GPT-4o, on the other hand, maintains more balanced performance across all dimensions.

\begin{table}[t]
\caption{MAE for scoring English and Math across urban and rural settings, calculated separately for multiple-choice (M), short-answer (S), essay (E), and total scores. The OCR results used in this analysis were obtained through \textbf{human transcription}. Lower values indicate better performance; bolded values represent the best results. Each component is scored on a 0–100 scale.}

\centering
\resizebox{0.95\linewidth}{!}{%
\begin{tabular}{@{}lcccclcccc@{}}
\toprule
    \multicolumn{1}{l}{}     & \multicolumn{4}{c}{\textbf{English}}          & \textbf{} & \multicolumn{4}{c}{\textbf{Math}} \\
    \cmidrule{2-5}\cmidrule{7-10}
    \multicolumn{1}{c}{\multirow{-2}{*}{\textbf{Model}}} & \multicolumn{1}{c}{\textbf{MCQ}}   & \multicolumn{1}{c}{\textbf{Short-Essay}}  & \multicolumn{1}{c}{\textbf{Essay}}   & \multicolumn{1}{c}{\textbf{Total}} & \textbf{} & \multicolumn{1}{c}{\textbf{MCQ}}   & \multicolumn{1}{c}{\textbf{Short-Essay}}  & \multicolumn{1}{c}{\textbf{Essay}}   & \multicolumn{1}{c}{\textbf{Total}} \\
    \midrule
    \rowcolor[HTML]{D9EAD3} 
    
    Urban & \multicolumn{1}{l}{\cellcolor[HTML]{D9EAD3}} & \multicolumn{1}{l}{\cellcolor[HTML]{D9EAD3}} & \multicolumn{1}{l}{\cellcolor[HTML]{D9EAD3}} & \multicolumn{1}{l}{\cellcolor[HTML]{D9EAD3}} & & \multicolumn{1}{l}{\cellcolor[HTML]{D9EAD3}} & \multicolumn{1}{l}{\cellcolor[HTML]{D9EAD3}} & \multicolumn{1}{l}{\cellcolor[HTML]{D9EAD3}} & \multicolumn{1}{l}{\cellcolor[HTML]{D9EAD3}} \\
    GPT4o  & \textbf{0.0}  & 2.4  & 7.2  & 0.8  && \textbf{0.0}  & 5.8  & \textbf{7.6}  & 2.4 \\
	Llama 3.1 (70B)  & \textbf{0.0}  & 7.7  & 2.9  & 2.7  && \textbf{0.0}  & 10.3  & 30.0  & 10.4 \\
	Qwen2.5 (72B)  & \textbf{0.0}  & 1.9  & \textbf{1.3}  & \textbf{0.5}  && \textbf{0.0}  & 7.6  & 10.7  & 3.9 \\
	Deepseek (671B)  & \textbf{0.0}  & \textbf{1.3}  & 3.5  & 1.5  && \textbf{0.0}  & \textbf{5.6}  & 9.9  & \textbf{1.0} \\
    \rowcolor[HTML]{D9EAD3} 
    Rural & \multicolumn{1}{l}{\cellcolor[HTML]{D9EAD3}} & \multicolumn{1}{l}{\cellcolor[HTML]{D9EAD3}} & \multicolumn{1}{l}{\cellcolor[HTML]{D9EAD3}} & \multicolumn{1}{l}{\cellcolor[HTML]{D9EAD3}} & & \multicolumn{1}{l}{\cellcolor[HTML]{D9EAD3}} & \multicolumn{1}{l}{\cellcolor[HTML]{D9EAD3}} & \multicolumn{1}{l}{\cellcolor[HTML]{D9EAD3}} & \multicolumn{1}{l}{\cellcolor[HTML]{D9EAD3}} \\
	GPT4o  & \textbf{0.0}  & 21.2  & \textbf{5.2}  & 23.1  && \textbf{0.0}  & 2.5  & \textbf{2.2}  & \textbf{0.3} \\
	Llama 3.1 (70B)  & \textbf{0.0}  & 26.1  & 11.4  & 26.9  && \textbf{0.0}  & 8.8  & 23.1  & 9.7 \\
	Qwen2.5 (72B)  & \textbf{0.0}  & 19.8  & 12.5  & 24.3  && \textbf{0.0}  & \textbf{1.7}  & 5.0  & 2.1 \\
	Deepseek (671B)  & \textbf{0.0}  & \textbf{14.2}  & 10.1  & \textbf{21.2}  && \textbf{0.0}  & 6.4  & 5.9  & 0.6 \\
    \bottomrule
\end{tabular}
}
\label{tab:result2}
\end{table}


\paragraph{Urban vs. Rural Performance Analysis} Given the significant educational disparities between rural and urban areas, we evaluated the performance of the model in these two settings. To isolate the analysis of LLM scoring capabilities, we use only the human-transcribed OCR results, eliminating recognition errors. 

Table~\ref{tab:result2} presents the MAE scores for English and Math, separated by question type: multiple choice (M), short answer (S), essay (E), and total scores. The results indicate that English MAEs are generally higher in rural settings than in urban settings across all models. For example, GPT-4o achieves a total MAE of only 0.8 in urban English, but this rises sharply to 23.1 in the rural setting. This discrepancy suggests that LLMs may struggle more in interpreting free-form responses from rural students, possibly due to variations in writing style and grammar. In contrast, MAEs for Math tend to be slightly lower in rural areas, although the differences are less pronounced. This may be attributed to the nature of Math questions, which often involve numerical reasoning and have more deterministic answers, reducing ambiguity in scoring.

\label{par:ocr_performance_analysis}
\paragraph{OCR Performance Analysis}
Given the differences in MAE between the GPT-4o OCR outputs and human transcription shown in Table~\ref{tab:result1}, we further analyze the OCR performance of GPT-4o and evaluate the extent to which recognition errors propagate to the subsequent scoring. For this analysis, we use exact string matching to assess accuracy on multiple-choice and short answer questions, and compute ROUGE-L \cite{lin-2004-rouge} scores to compare GPT-4o and human transcriptions for essay questions. 

Table~\ref{tab:result3} shows that the OCR performance is generally higher for English than for Math. Within English, responses from urban students yield higher exact match and ROUGE-L scores compared to those from rural students, possibly due to differences in handwriting clarity or writing conventions. For Math, the OCR accuracy is overall lower than that of English, but the performance gap between urban and rural settings is less pronounced. This suggests that while English responses may be more affected by region-specific handwriting variability, Math responses, often more structured and numerical, are comparatively stable across regions.

\begin{table}[t]
\caption{OCR-based performance (GPT-4o) across Urban, Rural, and All settings for English and Math: EM = exact match, RL = ROUGE-L F1, M = multiple choice, S = Short Essay, E = Essay.}
\centering
\resizebox{0.95\linewidth}{!}{%
\begin{tabular}{lcccccc}
\toprule
   & \multicolumn{3}{c}{\cellcolor[HTML]{FFF2CC}\textbf{English}} & \multicolumn{3}{c}{\cellcolor[HTML]{D9EAD3}\textbf{Math}}    \\
   \cmidrule{2-7}

\multirow{-2}{*}{\textbf{Area}} & \multicolumn{1}{c}{\cellcolor[HTML]{FFF2CC}\textbf{EM(MCQ)}} & \multicolumn{1}{c}{\cellcolor[HTML]{FFF2CC}\textbf{EM(Short-Essay)}} & \multicolumn{1}{c}{\cellcolor[HTML]{FFF2CC}\textbf{RL(Essay)}} & \multicolumn{1}{c}{\cellcolor[HTML]{D9EAD3}\textbf{EM(MCQ)}} & \multicolumn{1}{c}{\cellcolor[HTML]{D9EAD3}\textbf{EM(Short-Essay)}} & \multicolumn{1}{c}{\cellcolor[HTML]{D9EAD3}\textbf{RL(Essay)}} \\
\midrule
\cellcolor[HTML]{F8F8F8}{\color[HTML]{1D1C1D} Urban} & \textbf{82.1} & \textbf{67.1} & \textbf{60.3} & 62.3 & 23.3 & 21.0 \\
Rural      & 71.7 & 61.8 & 60.1 & \textbf{62.5} & \textbf{27.9} & \textbf{24.8} \\
\midrule
All        & 78.5 & 65.3 & 60.2 & 62.4 & 24.9 & 22.3     \\
\bottomrule
\end{tabular}
}

\label{tab:result3}
\end{table}

\section{Conclusion}
We release the first dataset of handwritten student answer sheets for AI-based assessment in Indonesia, covering over 14K Math and English responses from rural and urban schools. Using this dataset, we evaluate VLMs and LLMs for grading and personalized feedback, finding that GPT-4o and DeepSeek (671B) perform competitively with teacher scores, with DeepSeek producing more contextually relevant feedback. This work lays the foundation for AI-powered assessment in low-resource and underrepresented educational contexts.

\begin{credits}
\subsubsection{\ackname} This research was supported by the 2023 SEAMEO-Australia Education Links Award, jointly organized by the Southeast Asian Ministers of Education Organization (SEAMEO) Secretariat and the Australian Government Department of Education.

\subsubsection{\discintname}
The authors have no competing interests to declare that are relevant to the content of this article.
\subsubsection{Ethic Statement}
This study strictly adheres to ethical research practices in AI and education:
(i) All student answer sheets were anonymized prior to analysis. Identifying information, including names, school names, and class identifiers, was removed to protect student privacy and comply with ethical guidelines for research involving minors; (ii) Written informed consent was obtained from school administrators and participating teachers. Participation in the study was voluntary, and students were not penalized for opting out; (iii) The inclusion of both urban and rural schools was an intentional decision to ensure representation across socio-economic and educational divides. However, we recognize that the deployment of AI tools in such settings must be approached cautiously to avoid reinforcing existing inequalities. This study advocates for equitable development, localization, and participatory design of AI tools in education, particularly when applied in under-resourced areas.
\end{credits}

%
%
%
\bibliographystyle{splncs04}
\bibliography{reference}

@article{liu2024deepseek,
  title={Deepseek-v3 technical report},
  author={Liu, Aixin and Feng, Bei and Xue, Bing and Wang, Bingxuan and Wu, Bochao and Lu, Chengda and Zhao, Chenggang and Deng, Chengqi and Zhang, Chenyu and Ruan, Chong and others},
  journal={arXiv preprint arXiv:2412.19437},
  year={2024}
}

@article{daniela2025evaluating,
  title={Evaluating Handwritten Student Work in Latvian Using LLMs’ General Knowledge},
  author={Daniela, Linda and Norgaila, Edgaras and Kalni{\c{n}}a, Daiga},
  journal={Technology, Knowledge and Learning},
  pages={1--18},
  year={2025},
  publisher={Springer}
}

@inproceedings{lin-2004-rouge,
    title = "{ROUGE}: A Package for Automatic Evaluation of Summaries",
    author = "Lin, Chin-Yew",
    booktitle = "Text Summarization Branches Out",
    month = jul,
    year = "2004",
    address = "Barcelona, Spain",
    publisher = "Association for Computational Linguistics",
    url = "https://aclanthology.org/W04-1013/",
    pages = "74--81"
}

@article{steiner2024paligemma,
  title={Paligemma 2: A family of versatile vlms for transfer},
  author={Steiner, Andreas and Pinto, Andr{\'e} Susano and Tschannen, Michael and Keysers, Daniel and Wang, Xiao and Bitton, Yonatan and Gritsenko, Alexey and Minderer, Matthias and Sherbondy, Anthony and Long, Shangbang and others},
  journal={arXiv preprint arXiv:2412.03555},
  year={2024}
}

@article{lee2025realizing,
  title={Realizing visual question answering for education: GPT-4V as a multimodal AI},
  author={Lee, Gyeonggeon and Zhai, Xiaoming},
  journal={TechTrends},
  pages={1--17},
  year={2025},
  publisher={Springer}
}

@inproceedings{westerlund2024llm,
  title={LLM Integration in Workbook Design for Teaching Coding Subjects},
  author={Westerlund, Magnus and Shcherbakov, Andrey},
  booktitle={International Conference on Smart Technologies \& Education},
  pages={77--85},
  year={2024},
  organization={Springer}
}

@article{hu2025exploring,
  title={Exploring the potential of LLM to enhance teaching plans through teaching simulation},
  author={Hu, Bihao and Zhu, Jiayi and Pei, Yiying and Gu, Xiaoqing},
  journal={npj Science of Learning},
  volume={10},
  number={1},
  pages={7},
  year={2025},
  publisher={Nature Publishing Group UK London}
}

@article{zhang2024accessing,
  title={Accessing gpt-4 level mathematical olympiad solutions via monte carlo tree self-refine with llama-3 8b},
  author={Zhang, Di and Huang, Xiaoshui and Zhou, Dongzhan and Li, Yuqiang and Ouyang, Wanli},
  journal={arXiv preprint arXiv:2406.07394},
  year={2024}
}

@article{yeadon2024impact,
  title={The impact of AI in physics education: a comprehensive review from GCSE to university levels},
  author={Yeadon, Will and Hardy, Tom},
  journal={Physics Education},
  volume={59},
  number={2},
  pages={025010},
  year={2024},
  publisher={IOP Publishing}
}

@article{wei2022chain,
  title={Chain-of-thought prompting elicits reasoning in large language models},
  author={Wei, Jason and Wang, Xuezhi and Schuurmans, Dale and Bosma, Maarten and Xia, Fei and Chi, Ed and Le, Quoc V and Zhou, Denny and others},
  journal={Advances in neural information processing systems},
  volume={35},
  pages={24824--24837},
  year={2022}
}

@inproceedings{wang2023can,
  title={Can ChatGPT Defend its Belief in Truth? Evaluating LLM Reasoning via Debate},
  author={Wang, Boshi and Yue, Xiang and Sun, Huan},
  booktitle={Findings of the Association for Computational Linguistics: EMNLP 2023},
  pages={11865--11881},
  year={2023}
}

@article{liu2024vmamba,
  title={Vmamba: Visual state space model},
  author={Liu, Yue and Tian, Yunjie and Zhao, Yuzhong and Yu, Hongtian and Xie, Lingxi and Wang, Yaowei and Ye, Qixiang and Jiao, Jianbin and Liu, Yunfan},
  journal={Advances in neural information processing systems},
  volume={37},
  pages={103031--103063},
  year={2024}
}

@article{liu2023visual,
  title={Visual instruction tuning},
  author={Liu, Haotian and Li, Chunyuan and Wu, Qingyang and Lee, Yong Jae},
  journal={Advances in neural information processing systems},
  volume={36},
  pages={34892--34916},
  year={2023}
}

@inproceedings{morris2023using,
  title={Using Large Language Models to Provide Formative Feedback in Intelligent Textbooks},
  author={Morris, Wesley and Crossley, Scott and Holmes, Langdon and Ou, Chaohua and McNamara, Danielle and Dascalu, Mihai},
  booktitle={International Conference on Artificial Intelligence in Education},
  pages={484--489},
  year={2023},
  organization={Springer}
}

@article{touvron2023llama,
  title={Llama 2: Open foundation and fine-tuned chat models},
  author={Touvron, Hugo and Martin, Louis and Stone, Kevin and Albert, Peter and Almahairi, Amjad and Babaei, Yasmine and Bashlykov, Nikolay and Batra, Soumya and Bhargava, Prajjwal and Bhosale, Shruti and others},
  journal={arXiv preprint arXiv:2307.09288},
  year={2023}
}

@inproceedings{botarleanu2022multitask,
  title={Multitask summary scoring with longformers},
  author={Botarleanu, Robert-Mihai and Dascalu, Mihai and Allen, Laura K and Crossley, Scott Andrew and McNamara, Danielle S},
  booktitle={International Conference on Artificial Intelligence in Education},
  pages={756--761},
  year={2022},
  organization={Springer}
}

@misc{jiang2023mistral7b,
      title={Mistral 7B}, 
      author={Albert Q. Jiang and Alexandre Sablayrolles and Arthur Mensch and Chris Bamford and Devendra Singh Chaplot and Diego de las Casas and Florian Bressand and Gianna Lengyel and Guillaume Lample and Lucile Saulnier and Lélio Renard Lavaud and Marie-Anne Lachaux and Pierre Stock and Teven Le Scao and Thibaut Lavril and Thomas Wang and Timothée Lacroix and William El Sayed},
      year={2023},
      eprint={2310.06825},
      archivePrefix={arXiv},
      primaryClass={cs.CL},
      url={https://arxiv.org/abs/2310.06825}, 
}

@inproceedings{chiang2024large,
  title={Large Language Model as an Assignment Evaluator: Insights, Feedback, and Challenges in a 1000+ Student Course},
  author={Chiang, Cheng-Han and Chen, Wei-Chih and Kuan, Chun-Yi and Yang, Chienchou and Lee, Hung-Yi},
  booktitle={Proceedings of the 2024 Conference on Empirical Methods in Natural Language Processing},
  pages={2489--2513},
  year={2024}
}

@inproceedings{stahl2024exploring,
  title={Exploring LLM Prompting Strategies for Joint Essay Scoring and Feedback Generation},
  author={Stahl, Maja and Biermann, Leon and Nehring, Andreas and Wachsmuth, Henning},
  booktitle={Proceedings of the 19th Workshop on Innovative Use of NLP for Building Educational Applications (BEA 2024)},
  pages={283--298},
  year={2024}
}

@inproceedings{yancey2023rating,
  title={Rating short L2 essays on the CEFR scale with GPT-4},
  author={Yancey, Kevin P and Laflair, Geoffrey and Verardi, Anthony and Burstein, Jill},
  booktitle={Proceedings of the 18th workshop on innovative use of NLP for building educational applications (BEA 2023)},
  pages={576--584},
  year={2023}
}

@misc{touvron2023llamaopenefficientfoundation,
      title={LLaMA: Open and Efficient Foundation Language Models}, 
      author={Hugo Touvron and Thibaut Lavril and Gautier Izacard and Xavier Martinet and Marie-Anne Lachaux and Timothée Lacroix and Baptiste Rozière and Naman Goyal and Eric Hambro and Faisal Azhar and Aurelien Rodriguez and Armand Joulin and Edouard Grave and Guillaume Lample},
      year={2023},
      eprint={2302.13971},
      archivePrefix={arXiv},
      primaryClass={cs.CL},
      url={https://arxiv.org/abs/2302.13971}, 
}

@misc{qwen2.5,
    title = {Qwen2.5: A Party of Foundation Models},
    url = {https://qwenlm.github.io/blog/qwen2.5/},
    author = {Qwen Team},
    month = {September},
    year = {2024}
}

\end{document}